\documentclass[conference]{IEEEtran}
\usepackage{times}

\usepackage[numbers]{natbib}
\usepackage{multicol}
\usepackage{amsmath}

\def\etal{{\em et al.}}

\DeclareMathAlphabet\mathbfcal{OMS}{cmsy}{b}{n}
\def\0{{\bf 0}}
\def\1{{\bf 1}}

\def\etal{\emph{et al.}}

\usepackage{amsmath}

\usepackage{dsfont} 
\usepackage{booktabs}  
\usepackage{multirow}    
\usepackage{graphicx}     
\usepackage{amsmath}    
\usepackage{xspace}
\usepackage{subcaption}
\usepackage{lipsum}
\let\labelindent\relax
\usepackage{enumitem}

\usepackage{booktabs}
\usepackage{multirow}
\usepackage{array}
\usepackage[table]{xcolor}
\usepackage{duckuments}

\definecolor{LocoCol}{RGB}{220,235,245} %
\definecolor{ManiCol}{RGB}{245,225,225} %
\definecolor{LocoText}{RGB}{60,120,180}  
\definecolor{ManiText}{RGB}{180,70,70}   
\definecolor{HumanData}{RGB}{64,133,245}
\definecolor{RobotData}{RGB}{234,66,53}
\definecolor{LocoManip}{RGB}{50,168,82}

\usepackage{color}
\usepackage{colortbl}

\definecolor{deemph}{gray}{0.6}

\definecolor{baselinecolor}{gray}{.9}
\definecolor{yellow}{RGB}{218,165,32}
\definecolor{lightcyan}{rgb}{0.88, 1.0, 1.0}
\definecolor{lightskyblue}{rgb}{0.53, 0.81, 0.98}
\definecolor{aliceblue}{rgb}{0.94, 0.97, 1.0}
\definecolor{LightSlateBlue}{RGB}{70,130,180}
\definecolor{DeepBlue}{RGB}{65,100,170}
\definecolor{DeepPurple}{RGB}{136,105,160}
\definecolor{LightGreen}{RGB}{59,125,35}
\definecolor{LightRed}{RGB}{234,66,53}
\definecolor{cvprblue}{rgb}{0.21,0.49,0.74}

\usepackage{epigraph} 
\usepackage{booktabs}  
\usepackage[pagebackref,breaklinks,colorlinks,linkcolor=red,filecolor=magenta, citecolor=cvprblue]{hyperref}
\usepackage{url}
\usepackage{cleveref}
\crefname{section}{Sec.}{Secs.}
\Crefname{section}{Section}{Sections}
\Crefname{figure}{Figure}{Figures}
\crefname{figure}{Fig.}{Figs.}
\Crefname{table}{Table}{Tables}
\crefname{table}{Tab.}{Tabs.}

\usepackage{caption} %
\newcommand{\boldparagraph}[1]{\vspace{0.1cm}\noindent{\bf #1.}}
\newcommand{\modelname}{\textsc{EgoHumanoid}\xspace}

\begin{document}
\title{
Unlocking In-the-Wild Loco-Manipulation with Robot-Free Egocentric Demonstration
}

\author{\authorblockN{
Modi Shi$^{2,3*}$ \quad
Shijia Peng$^{1*\dagger}$ \quad
Jin Chen$^{2*}$ \quad
Haoran Jiang$^{2}$ \quad
Tianyu Li$^{2}$\\
Di Huang$^{3}$ \quad
Ping Luo$^{1\dagger}$ \quad
Hongyang Li$^{1}$$^{\natural}$ \quad
Li Chen$^{1}$$^{\natural}$\\
}
\vspace{0.1em}
\authorblockA{
$^{1}$ The University of Hong Kong  \quad
$^{2}$ Shanghai Innovation Institute \quad
$^{3}$ Beihang University \\
\vspace{0.1em}
\texttt{\url{https://opendrivelab.com/EgoHumanoid}}
}
}

\twocolumn[{%
	\renewcommand\twocolumn[1][]{#1}%
	\maketitle
	\begin{center}
		\includegraphics[width=.92\textwidth]{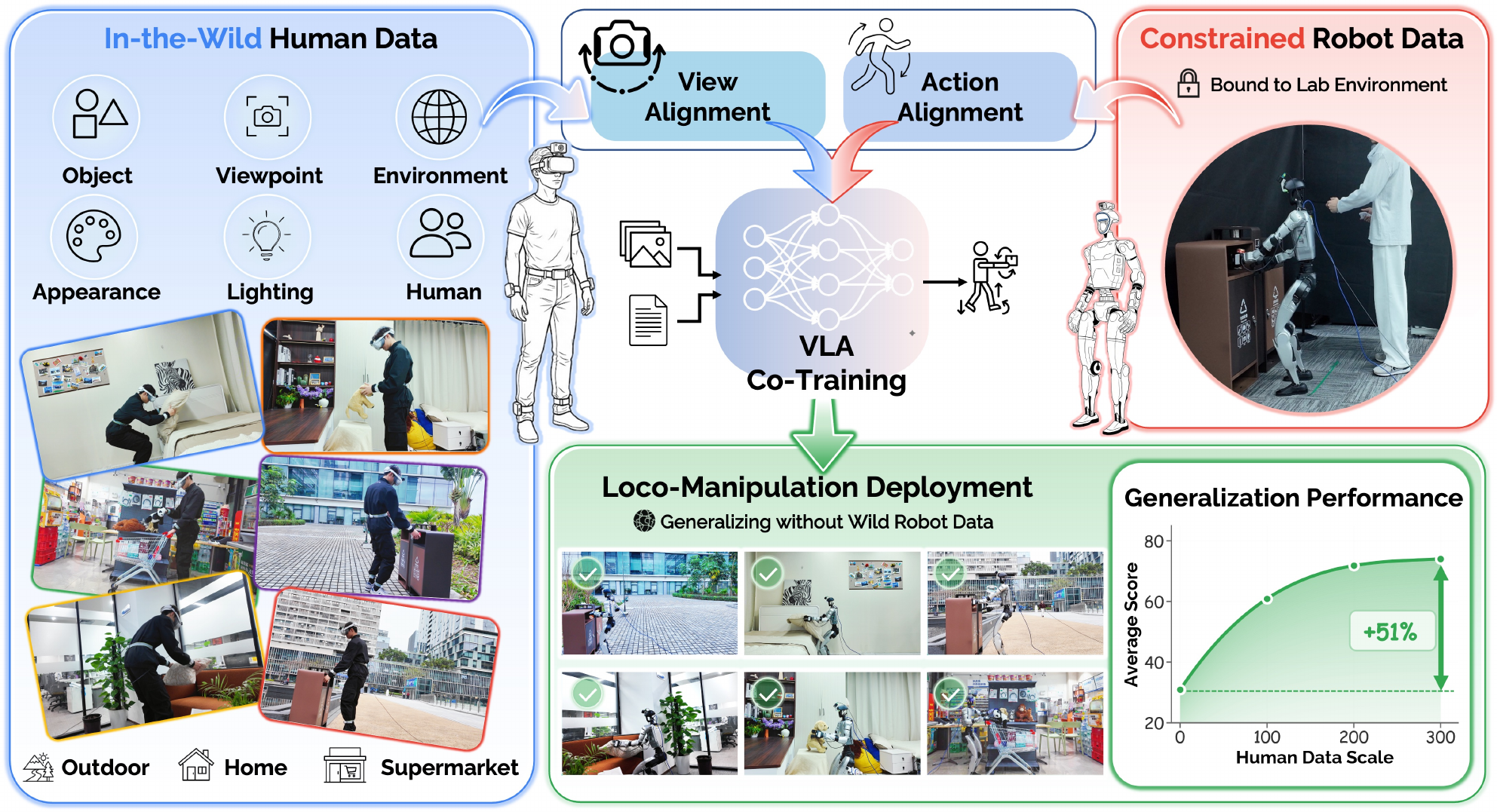}
		\captionof{figure}{
        Introducing \textbf{\modelname}, the first investigation
        on human-to-humanoid transfer for whole-body loco-manipulation. 
        \textcolor{RobotData}{Robot teleoperation data} collection is constrained to laboratory environment due to hardware and safety limitations,
        while \textcolor{HumanData}{in-the-wild human demonstrations} offer scalable diversity in objects, scenes, lighting, and viewpoints.
        Our alignment pipeline bridges the embodiment gap through view and action alignment, enabling vision-language-action (VLA)
        co-training on both data sources.
        \textcolor{LocoManip}{Real-world loco-manipulation deployment} validates that egocentric human demonstrations invigorate generalization \textit{without} scene-specific robot data,
        outperforming robot-only baselines by 51\% with consistent scaling behavior. 
        }
        \label{fig:teaser}
	\end{center}
}]

\begingroup
\renewcommand{\thefootnote}{}
\footnotetext{\hspace{-1.8em}$^{*}$Equal Contribution.\quad $^{\natural}$Project lead.\quad $^{\dagger}$Work done while at Kinetix AI.}
\begin{abstract}

Human demonstrations offer rich environmental diversity and scale naturally, making them an appealing alternative to robot teleoperation. While this paradigm has advanced robot-arm manipulation, its potential for the more challenging, data-hungry problem of humanoid loco-manipulation remains largely unexplored.
We present \modelname, the first framework to co-train a vision-language-action policy using abundant egocentric human demonstrations together with a limited amount of robot data, enabling humanoids to perform loco-manipulation across diverse real-world environments. To bridge the embodiment gap between humans and robots, including discrepancies in physical morphology and viewpoint, we introduce a systematic alignment pipeline spanning from hardware design to data processing. A portable system for scalable human data collection is developed, and we establish practical collection protocols to improve transferability. At the core of our human-to-humanoid alignment pipeline lies two key components. The view alignment reduces visual domain discrepancies caused by camera height and perspective variation. The action alignment maps human motions into a unified, kinematically feasible action space for humanoid control. Extensive real-world experiments demonstrate that incorporating robot-free egocentric data significantly outperforms robot-only baselines by 51\%, particularly in unseen environments. Our analysis further reveals which behaviors transfer effectively and the potential for scaling human data.
\end{abstract}

\IEEEpeerreviewmaketitle

\section{Introduction}
\label{sec:intro}

Humanoid robots hold immense promise for operating across diverse human-centric environments, from household assistance to outdoor service scenarios~\cite{gu2025humanoid}. These applications fundamentally demand \textit{loco-manipulation}, which involves the 
close orchestration
of whole-body locomotion and dexterous manipulation~\cite{lu2025mobiletelevision, ulc, falcon}. 
Unlike fixed-base manipulators~\cite{kim2024openvla, bu2025agibot_iros, sima2026kai0}, robots must navigate through varied spaces, adjust their body posture for reachability, and manipulate objects, all while maintaining dynamic balance in unstructured environments.

Despite rapid progress in model architectures and control algorithms, learning humanoid loco-manipulation remains bottlenecked by the \textit{scarcity} of diverse, large-scale demonstration data.
Existing approaches~\cite{fu2024mobilealoha, homie} primarily rely on robot teleoperation, which provides embodiment-consistent supervision but suffers from high cost, operational complexity, and hardware instability~\cite{lu2025mobiletelevision, he2025viral, twist}. 
Moreover, it confines data collection to the laboratory environment, as transporting humanoid platforms and teleoperation equipment such as motion capture suits to diverse real-world scenarios, \textit{e.g.}, homes, parks, stores, outdoor spaces, is often impractical~\cite{zhong2025humanoidexo, ze2025twist2}.

We instead provide a compelling alternative 
based on the simple observation: humans naturally perform loco-manipulation tasks every day, exactly across  the environments where robots are expected to operate. Recent advances in wearable sensing~\cite{engel2023project, grauman2024egoexo4d, robotics2025WIYH_TARS} make it possible to capture egocentric human demonstrations using lightweight, portable devices, without requiring any robot hardware. 
This paradigm offers scalable access to behaviorally rich and environmentally diverse data, and has induced progress in robot manipulation and navigation~\cite{kumar2019VMSR, kareer2025egomimic, tao2025dexwild, zhu2025emma, qiu2025hat, kareer2025emergence, chen2026intelligent}.

However, 
applying egocentric human demonstrations to humanoid control is far from straightforward due to the fundamental \textit{embodiment gap}. 
Humanoid robots differ substantially from humans
in morphology and kinematics, including limb proportions and joint limits~\cite{zhao2025resmimic, yang2025omniretarget}.
Egocentric visual observations also differ --
humans observe their own hands and body, whereas robots perceive metallic manipulators from distinct viewpoints~\cite{chi2024umi, zhong2025humanoidexo}.
Moreover, motion dynamics diverge, as human walking patterns, body sway, and balance strategies do not directly transfer to robots with different mass distributions and actuation constraints~\cite{qiu2025hat, zhu2025emma}.
Collectively, these discrepancies are 
pronounced in loco-manipulation, where whole-body movement amplifies viewpoint change and introduces 
substantial motion variation
in the egocentric observations.
To this end, 
we present \textbf{\modelname}, a systematic framework for human–humanoid co-training as described in \cref{fig:teaser}.
Our key insight is that while low-level actions are embodiment-specific, high-level behavioral structures (\textit{e.g.}, navigation routes, object approach strategies, and task decomposition) \textit{could} transfer reliably when observations and motions are properly aligned. Moreover, human demonstrations in diverse real-world settings expose policies to variations that robot-only data rarely covers, enabling stronger in-the-wild generalization.
Accordingly, we build a data collection system that captures both robot-free 
human demonstrations and teleoperated humanoid robot data. The portable human setup integrates a VR headset, body trackers for pose estimation, and an egocentric camera, enabling scalable collection in varied scenarios without robot hardware. Complementary robot data, collected via VR-based teleoperation, provides embodiment-accurate supervision for manipulation-intensive behaviors.

The embodiment alignment pipeline is proposed to convert human demonstrations into robot-compatible training signals. Our approach focuses on two core components. The view alignment reduces visual discrepancies by transforming human egocentric observations to approximate robot viewpoints using depth-based reprojection and inpainting.
The action alignment employs a unified action space shared by humans and robots, using delta end-effector poses for upper body control and discrete commands for locomotion.
Together, these mechanisms guarantee the VLA policy co-training 
on both data sources.

\textbf{\modelname} is validated on a Unitree G1 humanoid robot across four indoor and outdoor loco-manipulation tasks. 
Experiments demonstrate that incorporating human data significantly improves policy performance by 20\% on average and enables generalization to scenes not covered by the robot data, yielding a 51\% 
performance gain.
Through extensive ablation studies, we scrutinize the transfer mechanism for sub-tasks, validate the scaling effectiveness, and identify critical design choices in our alignment pipeline.

In summary, the contributions are:
    \textbf{(a) The first 
    endorsement of human-to-humanoid transfer for whole-body loco-manipulation tasks.} 
    We show that egocentric human data effectively enhances humanoid policy through co-training, establishing the feasibility of cross-embodiment transfer with an integrated data collection and training framework.
    \textbf{(b) A principled embodiment alignment pipeline.} 
    We propose practical strategies to bridge human-robot embodiment gaps, combining view alignment to mitigate visual disparities and action alignment to enforce kinematic feasibility.
    \textbf{(c) Comprehensive real-world evaluation and 
    analysis.} 
    We characterize which behaviors transfer effectively, how to scale human data, and exhibit that human data substantially boosts generalization beyond the robot-only training environment.

We hope this study could encourage broader exploration of egocentric human data as a scalable pathway toward generalizable humanoid control, and 
lay the groundwork for transferring human behaviors to complex loco-manipulation skills.

\section{Related Work}
\label{sec:related}

\subsection{Egocentric Human Data for Robots}
\label{sec:related-data}

Egocentric human data has emerged as a promising source for robot learning, offering scalable collection in diverse environments without robot hardware. Various systems enable robot-free data capture, including handheld grippers with mounted cameras~\cite{chi2024umi, zhaxizhuoma2025fastumi}, wearable glasses~\cite{engel2023project}, and VR/AR headsets such as Apple Vision Pro, Meta Quest, and PICO. Such data has been leveraged in multiple ways for embodied AI. One line of work trains vision-language models on large-scale egocentric datasets~\cite{grauman2022ego4d, grauman2024egoexo4d, ye2025mmego, luo2025openmmego, lin2022egocentric, zhang2026sparsevideonav} to enhance scene understanding and spatial reasoning, with applications in embodied question answering~\cite{Majumdar2024OpenEQA} and egocentric decision-making~\cite{xu2025egodtm}. For robot policy learning, prior work primarily adopts a pretraining-then-finetuning paradigm, using human data to learn visual representations~\cite{nair2022r3m, ma2022vip, mpi}, motion priors~\cite{niu2025human2locoman, zhou2025mitigating, yang2025egovla, yu2025egomi}, or latent action embeddings~\cite{ye2024lapa, univla, jiang2025wholebodyvla}, discarding action information during pretraining. Co-training, in contrast, leverages actions from both human and robot demonstrations as supervision through aligned observation and action spaces, enabling more effective knowledge transfer~\cite{qiu2025hat, kareer2025emergence, zhu2025emma, yuan2025motiontrans}. However, existing co-training approaches focus on fixed-base manipulation. Our work presents the first validation of human-robot co-training for humanoid loco-manipulation, addressing the unique challenges of transferring whole-body human demonstrations to mobile humanoid platforms.

\subsection{Cross-Embodiment Data Alignment}
\label{sec:related-alignment}

In robot-to-robot transfer, Mirage~\cite{chen2024mirage} reduces visual mismatch by rewriting robots in images through cross-painting and inpainting, while Chen~\etal~\cite{chen2024rovi} use generative models to augment robot appearances and viewpoints.
For action alignment, common approaches either adopt transferable task-space or end-effector control interfaces~\cite{chen2024mirage, wei2024DRO-Grasp, shi2025diversity}, or project heterogeneous state–action spaces into a shared latent representation~\cite{wang2024cross, liu2024udh, dastider2025cross}.
Human-to-robot transfer is more challenging due to large gaps in morphology, viewpoint, and motion dynamics, but it is attractive given the scale of human data. The UMI series~\cite{chi2024umi, zhaxizhuoma2025fastumi, xu2025dexumi, wu2025freetacman} makes human demonstrations robot-compatible at collection time using hand-held hardware that provides accurate global positions. However, this hand-held hardware ties the collected data to a specific gripper morphology.
Another line of work relies on motion retargeting, mapping human motions to robot configurations~\cite{joao2025gmr, yang2025omniretarget, humanplus}, which is usually employed for training low-level humanoid controllers~\cite{omnih2o, gmt, beyondmimic}. This paradigm is further used by several concurrent manipulation or navigation-centered works, through estimating hand poses from egocentric human videos for joint training with robot data~\cite{kareer2025egomimic, zhu2025emma, kareer2025emergence, yuan2025motiontrans}. They focus on manipulation solely or decouple these two sub-tasks for mobile robots to prevent base shift. 
Contrarily, our work targets humanoid loco-manipulation with tightly-coupled whole-body coordination. We address the larger morphological 
gap through dedicated view transformation, including reprojection and generation, and action retargeting to fully exploit egocentric human demonstrations.

\subsection{Humanoid Loco-Manipulation}
\label{sec:related-locomanip}

Humanoid loco-manipulation couples locomotion and manipulation, as well as task planning, where both reliable execution and scalable supervision remain challenging.  Note that ``whole-body'' here is used at the task level rather than full joint-level actuation.
Inspired by humanoid whole-body control~\cite{omnih2o, videomimic, beyondmimic, pan2025ams}, emerging works build loco-manipulation skills with reinforcement learning, typically relying on specifically modeled objects in simulation, carefully designed rewards, and learning curricula to handle the strong coupling between base motion and arm–hand interaction~\cite{he2025viral, falcon, yin2025visualmimic, zhao2025resmimic}.  However, these methods face scaling bottlenecks such as per-contact physics tuning, intractable affordance modeling, and unmodeled hardware non-idealities.
In parallel, recent works explore scaling supervision through motion- or language-conditioned generation, aiming to lower the burden of expensive robot teleoperation needed for new tasks~\cite{fu2025demohlm, ding2025humanoidvla, xue2025leverb, taouil2025physically}. VLA policies have been developed
from manipulation methods as well, by extending models with lower-body robot action commands~\cite{jiang2025wholebodyvla, zhong2025humanoidexo, gr00t1.6} or motion targets~\cite{luo2025sonic, boston, helix02}. Despite this progress, most pipelines still scale primarily through robot-centered data or constrained setups, enduring the lack of environmental diversity and laborious data collection process, which prevents studying generalization in human-centric scenes. In sharp contrast, we treat robot-free egocentric human demonstrations as a scalable source of diverse whole-body behaviors to complement robot data, and we provide a starting point to co-train a humanoid VLA with two sources of data corpus through explicit view and action alignment.

\section{\modelname Framework}
\label{sec:method}

We present the \modelname~framework for training humanoid loco-manipulation policies by leveraging both egocentric human demonstrations and teleoperated robot data. The overview is illustrated in \cref{fig:teaser}. We first formalize the problem setup (\cref{sec:method-setup}), then describe our data collection system (\cref{sec:method-data}), embodiment alignment pipeline (\cref{sec:method-alignment}), and policy training procedure (\cref{sec:method-training}).

\subsection{Problem Setup}
\label{sec:method-setup}

Our goal is to train a VLA model capable of completing loco-manipulation tasks across novel real-world environments. The policy is trained on a combined dataset $\mathcal{D} = \mathcal{D}_\text{robot} \cup \mathcal{D}_\text{human}$, where $\mathcal{D}_\text{robot}$ consists of teleoperated robot demonstrations collected in constrained laboratory settings, and $\mathcal{D}_\text{human}$ comprises egocentric human demonstrations of the same tasks captured across diverse scenarios, \textit{e.g.}, homes, stores, and outdoor environments. Each data episode in $\mathcal{D}$ contains egocentric videos and synchronized whole-body actions.

After training, the policy is deployed under two settings: (1) \textit{in-domain} evaluation in laboratory environments similar to those in $\mathcal{D}_\text{robot}$, and (2) \textit{generalization} evaluation in scenes covered by $\mathcal{D}_\text{human}$ but absent from $\mathcal{D}_\text{robot}$. This setup directly tests whether human data enables generalization beyond the limited scenes where robot teleoperation is feasible.

We use a mid-sized (1.3m) Unitree G1 humanoid robot as our hardware platform, on account of its robust performance in low-level controller and durability. Meanwhile, the robot bares huge embodiment gap with normal teleoperators considering its size and degree of freedom (29 DoF with 3-finger Dex3 dexterous hands). Unlike prior work on tabletop bimanual manipulation, where the robot base remains stationary, we focus on tasks that require whole-body coordination—the robot must locomote to target locations while simultaneously or sequentially performing manipulation. This fundamental requirement for lower-body mobility distinguishes humanoid loco-manipulation from fixed-base settings and motivates our investigation of human-to-humanoid transfer.

\begin{figure}[t!]
    \centering
    \includegraphics[width=1\linewidth]{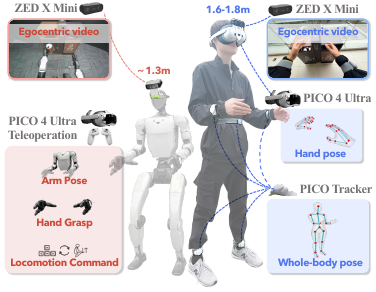}
    \caption{\textbf{Hardware setup for data collection.}
    Humans and the 
    G1 humanoid robot are equipped with an integrated VR-based system for portable usage and agile development. The same camera captures egocentric recordings. The VR headset and trackers provide coarse human poses, while the VR controller is employed to teleoperate the robot.
    } 
    \label{fig:hardware}
\end{figure}

\begin{figure*}[t!]
    \centering
    \includegraphics[width=.825\linewidth]{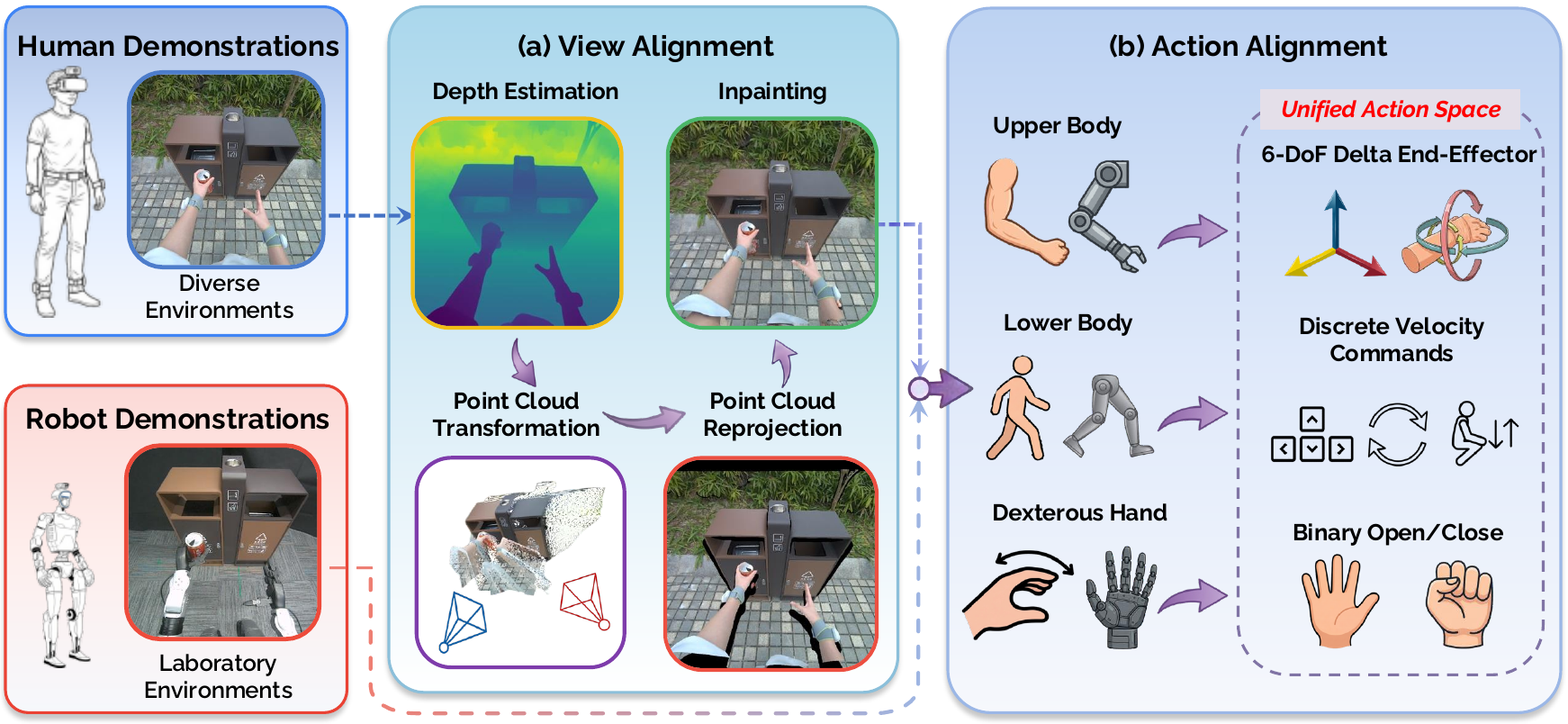}
    \caption{\textbf{Pipeline of human-to-humanoid alignment.}
    \textbf{(a) View Alignment}: Egocentric images are transformed to approximate robot viewpoints by reprojecting estimated depth points and generative inpainting to fill in blank holes.
    \textbf{(b) Action Alignment}: We employ relative end-effector poses to unify the upper body action space, and discrete commands for lower-body locomotion.
    }
    \label{fig:framework}
\end{figure*}

\subsection{Data Collection System}
\label{sec:method-data}

Crucially, robots and humans differ in practicality and embodiment. Robot data collection typically requires expensive hardware, careful setup, and is mostly bound to laboratories, while human data can be collected cheaply and flexibly using wearable devices in diverse settings. To mitigate the embodiment and visual gap in the data collection level and support human-robot co-training, we develop a unified and portable hardware setup with a VR-based system for swift adaptation between the two collection modes, shown in \cref{fig:hardware}. We intentionally drop wrist cameras as egocentric-only is a more universal setup, and the benefits of wrist views are not deterministic due to the enormous gaps, as shown in previous works~\cite{kareer2025emergence} (See more discussion in \cref{sec:discussion}). The efficiency of human demonstration contrast to robot teleoperation is pronounced ($\sim2\times$) and detailed in Table~\ref{tab:collection_time}.

\boldparagraph{Human Data Collection}
We use a portable PICO VR setup that enables low-cost and unconstrained recording of human demonstrations in both indoor and outdoor settings. The subject wears the headset with five PICO Motion Trackers, while a head-mounted ZED X Mini camera captures synchronized egocentric RGB images. Using the PICO SDK~\cite{zhao2025xrobotoolkit}, we record full-body human motion in real time, including 24 body keypoints and detailed hand poses with 26 keypoints per hand. 

\boldparagraph{Robot Data Collection}
Robot demonstrations are collected through VR-based teleoperation. The operator wears a PICO VR headset and uses handheld controllers to issue navigation commands (forward/backward, lateral movement, turning, standing, squatting) and wrist pose commands derived from the controller-to-headset relative pose. These commands are converted to joint-level actions through inverse kinematics and executed on a Unitree G1 humanoid, while the controller trigger controls grasping of a Dex3 dexterous hand. A low-level locomotion policy~\cite{GR00T-WholeBodyControl} ensures stable whole-body execution. We record navigation commands, wrist poses of the end-effector, hand grasp states, and synchronized egocentric RGB images from a head-mounted ZED X Mini camera.

\subsection{Human-to-Humanoid Alignment}
\label{sec:method-alignment}

To transform human demonstrations captured by the VR setup in \cref{sec:method-data} into robot-compatible training data, we develop an alignment pipeline consisting of two main modules, view alignment and action alignment for addressing visual domain gaps from differing camera viewpoints and action representation gaps from morphological discrepancies, respectively. The procedure is depicted in \cref{fig:framework} and described below. More implementation details are provided in Appendix~\ref{sec:supp-implementation}.

\boldparagraph{View Alignment} 
The substantial height discrepancy between humans and humanoid robots results in pronounced differences in egocentric visual observations. To bridge this visual gap, we transform human egocentric images to approximate robot camera viewpoints through a three-stage process. First, we use MoGe~\cite{wang2025moge} to infer an affine-invariant per-pixel 3D point map via its reprojection-based focal/shift recovery, and derive a scale-invariant depth map.
We then transform the recovered 3D points into the target robot camera frame and project them onto the target image plane. During training, random perturbations are applied to the target pose to encourage robustness to viewpoint variations. 
Note that the reprojection can produce missing regions due to invalid 3D predictions indicated by MoGe's validity mask and view-dependent disocclusions introduced by the pose change, leaving target pixels with no source correspondence. Therefore, the final step involves employing latent diffusion-based inpainting~\cite{Rombach_2022_CVPR} to hallucinate the missing regions, conditioning on the observed context and the missing region mask. Throughout this, complete RGB images are generated for better mimicking robot egocentric inputs.

\boldparagraph{Action Alignment} 
We design a unified action space that accommodates both human and robot demonstrations while respecting their kinematic differences.

\textit{Upper Body.} The actions are parameterized as 6-DoF delta end-effector poses, consistent with concurrent cross-embodiment works~\cite{zhu2025emma, kareer2025emergence, yuan2025motiontrans, yu2025egomi}. Using deltas avoids dependence on a globally aligned base frame, which could be ill-defined between human and robot recordings, enabling direct comparability across embodiments. We also avoid joint-level retargeting~\cite{joao2025gmr} as it may introduce artifacts and perturb interaction-relevant hand–object geometry~\cite{yang2025omniretarget}. Explicitly, we express human wrist poses in a pelvis-centric frame, 
then smooth translations with a Savitzky-Golay filter~\cite{SGFilter}. For rotations, we filter in the $\mathrm{SO}(3)$ tangent space using log/exp maps to prevent quaternion interpolation ambiguities. Finally, the transformed data is downsampled from 100\,Hz to 20\,Hz to match robot control and output the delta pose between consecutive frames as the action.

\textit{Lower body.} To collect robot data, the operator issues navigation commands using a discrete set of constant-velocity primitives (\textit{i.e.}, forward/backward, left/right, turn left/right, stand/squat) through the VR setup, following common practice in humanoid loco-manipulation~\cite{homie, jiang2025wholebodyvla}. To align human demonstrations with this robot action space, we convert the human pelvis trajectory into the same discrete commands. Specifically, we apply Savitzky–Golay smoothing to suppress jitter, then estimate the instantaneous heading via centered differences with a continuity constraint to prevent direction flips. 
Displacements in the world frame are projected on the local frame to obtain forward and lateral velocities. 
Yaw rate is computed from inter-frame heading changes. We downsample these continuous commands to 20\,Hz by averaging within each control window, and quantize them into discrete bins of horizontal movements.
Finally, a discrete stand/squat primitive is derived by thresholding inter-frame changes in pelvis height, matching the robot teleoperation interface.

\textit{Gripper.}
We represent the gripper action as a binary variable $a_t=\{0,1\}$, where $1$ corresponds to the gripper being closed and $0$ corresponds to the gripper being open. During robot data collection, a teleoperator controls the gripper via a handheld controller, and the resulting binary command sequence $\{a_t\}$ is recorded. Human demonstrations are captured using the VR system that tracks 26 hand joints per hand.
To robustly infer the hand grasping state, we first apply low-pass filtering followed by Savitzky-Golay smoothing
to the raw joint trajectories. We then compute the finger-level curvature $\kappa_f$, defined as the curvature evaluated at the midpoint of a quadratic polynomial fitted to each finger’s joint polyline. The hand-level grasping state is obtained by averaging $\kappa_f$ across all fingers and thresholding the resulting scalar $\bar{\kappa}$ to produce a binary open/close label. This curvature-based representation mitigates noise and enables reliable supervision extraction from human demonstrations. Similarly, all human-derived signals are downsampled to 20~Hz to align temporally with the robot data.

\subsection{Loco-Manipulation Policy Co-Training}
\label{sec:method-training}

With identical visual observation formats and action dimensions owing to the alignment procedure, we co-train a single policy on both human and robot demonstrations. Explicitly, we adapt $\pi_{0.5}$~\cite{intelligence2025pi05}, a state-of-the-art VLA model, for humanoid loco-manipulation via finetuning. The policy receives egocentric RGB observations and language instructions as input, and outputs actions in our unified action space. We intentionally omit proprioceptive states, as morphological differences between humans and humanoids yield incompatible proprioceptive distributions that would confound co-training.

\boldparagraph{Multi-source Data Sampling} In this case, the amount of human demonstrations may be substantially larger and more diverse than robot data, creating a highly imbalanced dataset. Prior work has shown that balanced sampling is essential for imbalanced multi-source training in robot manipulation~\cite{team2024octo, kalashnikov2021mt}. Meanwhile, loco-manipulation tasks require both high-level navigation or locomotion and intricate manipulation, where human and robot data may feature distinct advantages in these aspects. Thereafter, we examine the human-to-robot sampling ratio within each mini-batch, 
preserving sufficient exposure to robot demonstrations (See \cref{sec:exp-scaling}).

\begin{figure*}[t!]
    \centering
    \hspace{-0.5cm}
    \includegraphics[width=.9\linewidth]{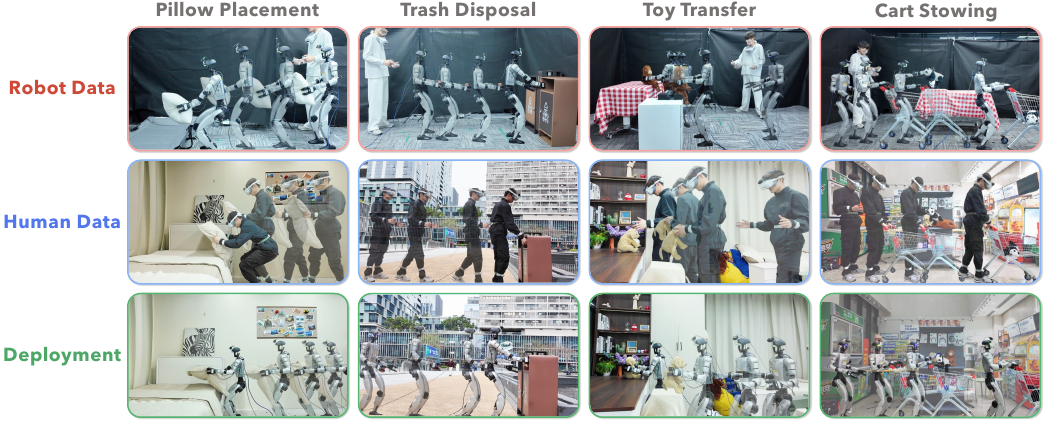}
    \caption{\textbf{Humanoid loco-manipulation tasks for evaluation.}
    We design four tasks which span varying levels of difficulty for large-space movement and dexterous manipulation.
    Robots are teleoperated in laboratories as the source domain (\textbf{Top}).
    Human-centric scenes 
    occur
    in human demonstrations only \textbf{(Middle)}, and set as testbeds for generalization evaluation (\textbf{Bottom}).
    }
    \label{fig:tasks}
\end{figure*}

\begin{figure*}[t!]
    \centering
    \includegraphics[width=.9\linewidth]{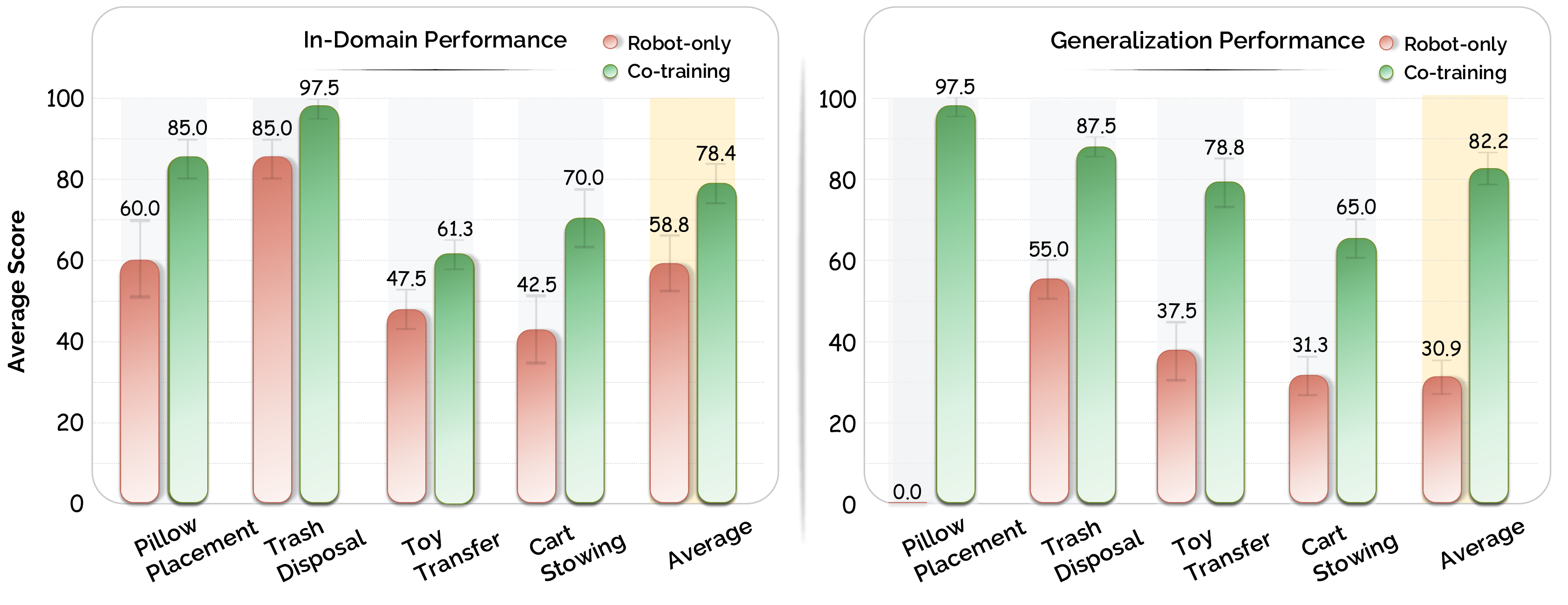}
    \caption{\textbf{Performance of human-robot data co-training with \modelname.}
    Our pipeline achieves unanimous improvements over robot-only baselines across in-domain and generalized environments. The boost is 
    amplified in robots' unseen settings. 
    }
    \label{fig:histogram}
\end{figure*}

\section{Evaluations}
\label{sec:exp}

We conduct real-world experiments to evaluate the effectiveness of leveraging egocentric human data for humanoid loco-manipulation. Our evaluation addresses three key questions:

\begin{itemize} [leftmargin=*]
    \item \textbf{Q1: Generalization.} Does egocentric human data enable humanoid policies to generalize to diverse human-centric environments beyond laboratory settings?
    
    \item \textbf{Q2: Transfer Analysis.} What behaviors transfer effectively from human demonstrations to humanoid policies?
    
    \item \textbf{Q3: Data Scaling.} Does policy performance scale with increasing amounts of human demonstration data?

\end{itemize}

\subsection{Real-world Experimental Setup}
\label{sec:exp-setup}

We design four loco-manipulation tasks requiring tight coordination between locomotion and manipulation, illustrated in \cref{fig:tasks}. All tasks involve non-trivial locomotion (1--5$m$), distinguishing 
them from fixed-base bimanual manipulation. Navigation accuracy 
subsequently
impacts manipulation success: suboptimal stopping positions render subsequent manipulation infeasible. 
The variance in stopping positions across trials further requires policies to generalize across viewpoints and object configurations.

The four tasks are: (1)
\noindent\texttt{Pillow Placement.} The robot carries a pillow to a bed and places it at a target location. This task tests stable locomotion with a bulky object and placement on a deformable surface.
(2)
\noindent\texttt{Trash Disposal.} The robot carries garbage (crumpled paper or a can) to a waste container with lids, inserts it horizontally into the hole rather than dropping from above, requiring precise localization and end-effector control.
(3)
\noindent\texttt{Toy Transfer.} The robot approaches a toy on a surface, grasps it, reorients, walks to a distant table, and places the toy down. This task evaluates sequential coordination across approach, grasp, in-hand locomotion, and placement.
(4)
\noindent\texttt{Cart Stowing.} The robot pushes a cart to a product display, grasps a toy, places it in the cart, and pushes the cart away. This task involves sustained contact during locomotion and multi-phase manipulation.

We conduct 20 trials per setting with position perturbations, and normalized scores are adopted to evaluate the performance of each trial. We delineate complementary experimental details in Appendix~\ref{sec:supp-exp-detail}, and list our scoring criteria in Appendix~\ref{sec:supp-score}.

\begin{table*}[t!]
\centering
\caption{
\textbf{Granular performance across tasks and subtasks.}
Each task is roughly categorized into sub-steps of locomotion (including high-level navigation) or manipulation, represented in \textcolor{LocoText}{blue} and \textcolor{ManiText}{red}, respectively, and denoted as s1, s2, \textit{etc}. 
}
\label{tab:task_subtask_data_setting}
\begin{tabular}{l >{\centering\arraybackslash}p{0.05\linewidth}>{\centering\arraybackslash}p{0.05\linewidth}>{\centering\arraybackslash}p{0.05\linewidth}>{\centering\arraybackslash}p{0.05\linewidth}>{\centering\arraybackslash}p{0.04\linewidth}>{\centering\arraybackslash}p{0.04\linewidth}>{\centering\arraybackslash}p{0.04\linewidth}>{\centering\arraybackslash}p{0.04\linewidth}>{\centering\arraybackslash}p{0.04\linewidth}>{\centering\arraybackslash}p{0.04\linewidth}>{\centering\arraybackslash}p{0.04\linewidth}>{\centering\arraybackslash}p{0.04\linewidth}}
\toprule
\multirow{2}{*}{Training Data}
 & \multicolumn{2}{c}{\textbf{Pillow Placement}} 
 & \multicolumn{2}{c}{\textbf{Trash Disposal}} 
 & \multicolumn{4}{c}{\textbf{Toy Transfer}} 
 & \multicolumn{4}{c}{\textbf{Cart Stowing}}  \\
\cmidrule(lr){2-3}
\cmidrule(lr){4-5}
\cmidrule(lr){6-9}
\cmidrule(lr){10-13}

 & \textcolor{LocoText}{s1}
 & \textcolor{ManiText}{s2} & \textcolor{LocoText}{s1} & \textcolor{ManiText}{s2}
 & \textcolor{LocoText}{s1} & \textcolor{ManiText}{s2} & \textcolor{LocoText}{s3}
 & \textcolor{ManiText}{s4} & \textcolor{LocoText}{s1} & \textcolor{ManiText}{s2} & \textcolor{ManiText}{s3} & \textcolor{LocoText}{s4} \\
\midrule

Robot-only   & 0   & 0  & 65 & 45 & 100 & 50 & 0 & 0 & 100 & 15 & 5 & 5 \\
Human-only   & 100 & 95 & 100 & 80 & 100 & 100 & 45 & 35 & 100 & 5 & 0 & 0 \\
Co-training  & \textbf{100} & \textbf{95} & \textbf{100} & \textbf{75} & \textbf{100} & \textbf{100} & \textbf{60} & \textbf{55} & \textbf{100} & \textbf{60} & \textbf{50} & \textbf{50} \\ 

\bottomrule
\end{tabular}
\end{table*}

\begin{figure*}[t!]
    \centering
    \includegraphics[width=.98\linewidth]{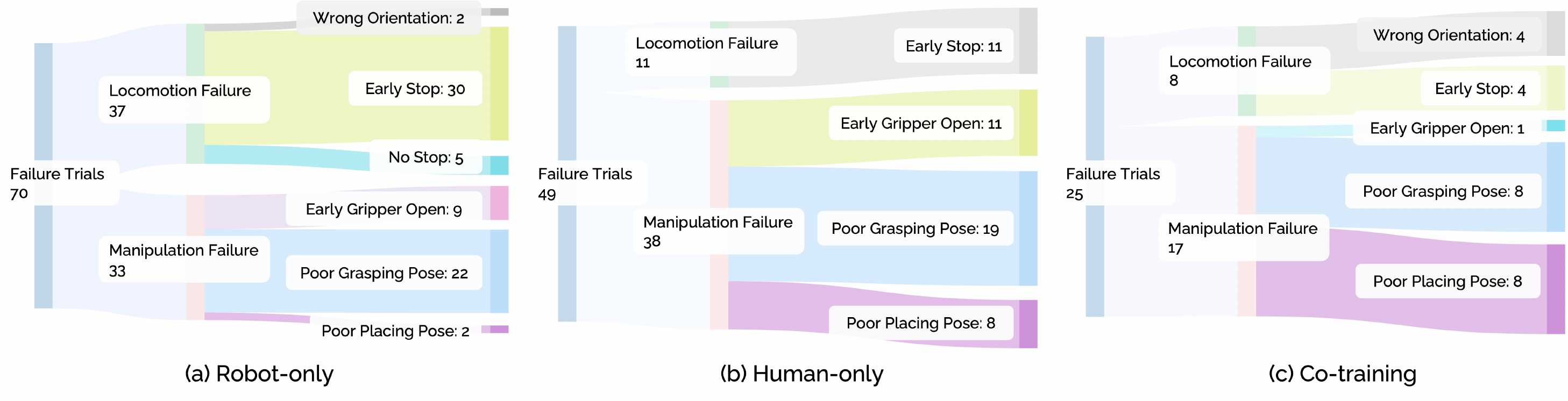}
    \caption{\textbf{Failure mode analysis.}
    For each setting—(a) Robot-only, (b) Human-only, and (c) Co-training—we log all failed episodes. The Sankey diagrams break failures down into locomotion vs.\ manipulation errors and finer causes.
    }
    \label{fig:failure}
\end{figure*}

\subsection{Will human data improve 
humanoid 
loco-manipulation?}
\label{sec:exp-main}

We evaluate whether co-training with robot-free egocentric demonstrations enables humanoid policies to generalize beyond laboratory environments. 
We compare \textit{Robot-only}, trained exclusively on robot data, with \textit{Co-training}, trained on both sources using our alignment pipeline. Policies are evaluated under two settings: \textit{In-Domain} (laboratory environments) and \textit{Generalization} (diverse scenes covered only by human data). For these experiments, robot-only models use 100 episodes per task, while co-training absorbs 300 episodes of human demonstrations additionally. 

As shown in \cref{fig:histogram}, co-training consistently improves performance across both settings. 
For in-domain evaluation, co-training achieves 78\% average score versus 59\% for robot-only. The gap widens remarkably in generalization settings: co-training reaches 82\% compared to 31\% for robot-only. Notably, generalization performance exceeds in-domain results, indicating that human data effectively bridges the domain gap to novel environments. These results demonstrate that humanoid robots can operate in diverse real-world settings without in-the-wild robot data collection.

\subsection{Which subskill benefits most from the data transfer?}
\label{sec:exp-subtask}
Delving deeper into the results in \cref{fig:histogram}, we examine 
which skills are primarily learned
from human demos. We train a \textit{Human-only} model and analyze performance separately across locomotion and manipulation stages. \Cref{tab:task_subtask_data_setting} shows consecutive success rates of sub-steps.
Though dependence exists among steps leading to a not strictly aligned `apple-to-apple' comparison, two observations emerge.

First, navigation transfers effectively from human data alone. Human-only achieves 100\% on navigation-dominated subtasks (Pillow Placement s1, Trash Disposal s1, Toy Transfer s1, Cart Stowing s1). Even on Toy Transfer s3, requiring two consecutive turns, Human-only attains 45\%, only 15\% below Co-training. In contrast, Robot-only yields near-zero success on these stages, confirming that human demonstrations provide transferable navigation knowledge.

Second, manipulation also transfers but effectiveness diminishes with increasing precision requirements. Human-only outperforms Robot-only on coarse manipulation subtasks (Pillow Placement s2, Trash Disposal s2, Toy Transfer s2/s4), where approximate contact suffices. However, for precision-critical phases such as Cart Stowing s2, Human-only achieves only 5\% versus Robot-only's 15\%. Notably, Co-training reaches 60\% on this subtask, indicating that human data provides useful priors that become effective when combined with robot demos.

The failure analysis in \cref{fig:failure} corroborates these findings: Human-only exhibits three times more manipulation failures than locomotion failures, whereas Robot-only failures are balanced across both phases.

\begin{figure*}[t!]
    \centering
    \includegraphics[width=0.98\linewidth]{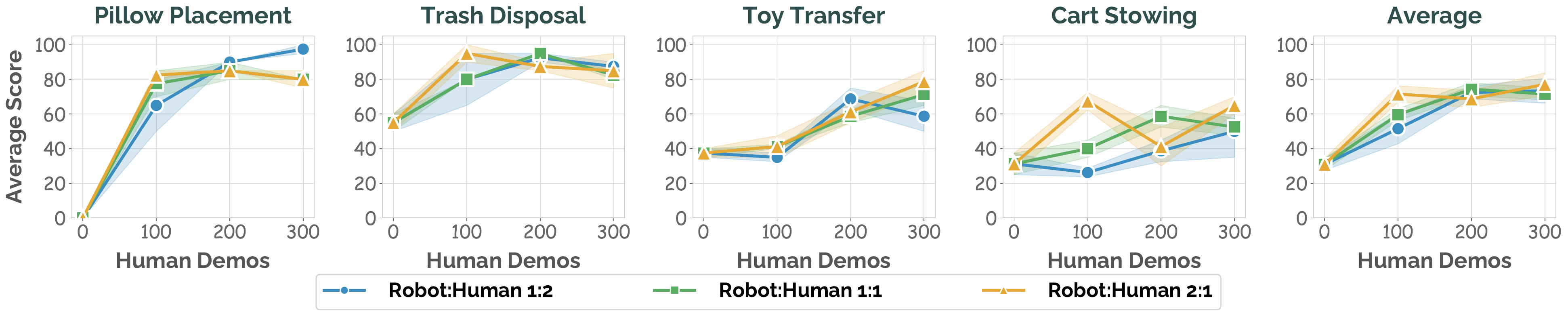}
    \caption{\textbf{Scaling human demonstrations with various training-time data sampling strategies.}
    More human demonstrations, from robot-only (0 Human Demos) to 3$\times$ robot data (300 Human Demos), principally yield improved loco-manipulation performance in generalized scenarios. The optimal sampling ratio depends on task characteristics: manipulation-heavy tasks favor higher robot data ratios, while navigation-dominant tasks benefit from more human data.
    } 
    \label{fig:scaling}
\end{figure*}

\subsection{Does Performance Scale with Human Data?}
\label{sec:exp-scaling}

We investigate how human demonstration quantity and mini-batch sampling ratio jointly affect policy performance. \cref{fig:scaling} presents results across four tasks as human demonstrations scale from 0 to 300 under varying sampling ratios. The optimal ratio depends on task characteristics: tasks requiring only coarse grasping (\textit{e.g.}, Pillow Delivery) benefit from higher human data ratios (Robot:Human 1:2), while fine manipulation tasks (\textit{e.g.}, Toy Transfer, Cart Shopping) favor higher robot ratios (Robot:Human 2:1). This aligns with our subtask analysis (\cref{sec:exp-subtask}), confirming that the embodiment gap disproportionately impacts precision-critical manipulation.

Across all sampling ratios, results improves consistently as human demonstrations accumulate, verifying that our method scales effectively without overfitting to dominant human data.

\begin{figure}[t!]
    \centering
    \includegraphics[width=.9\linewidth]{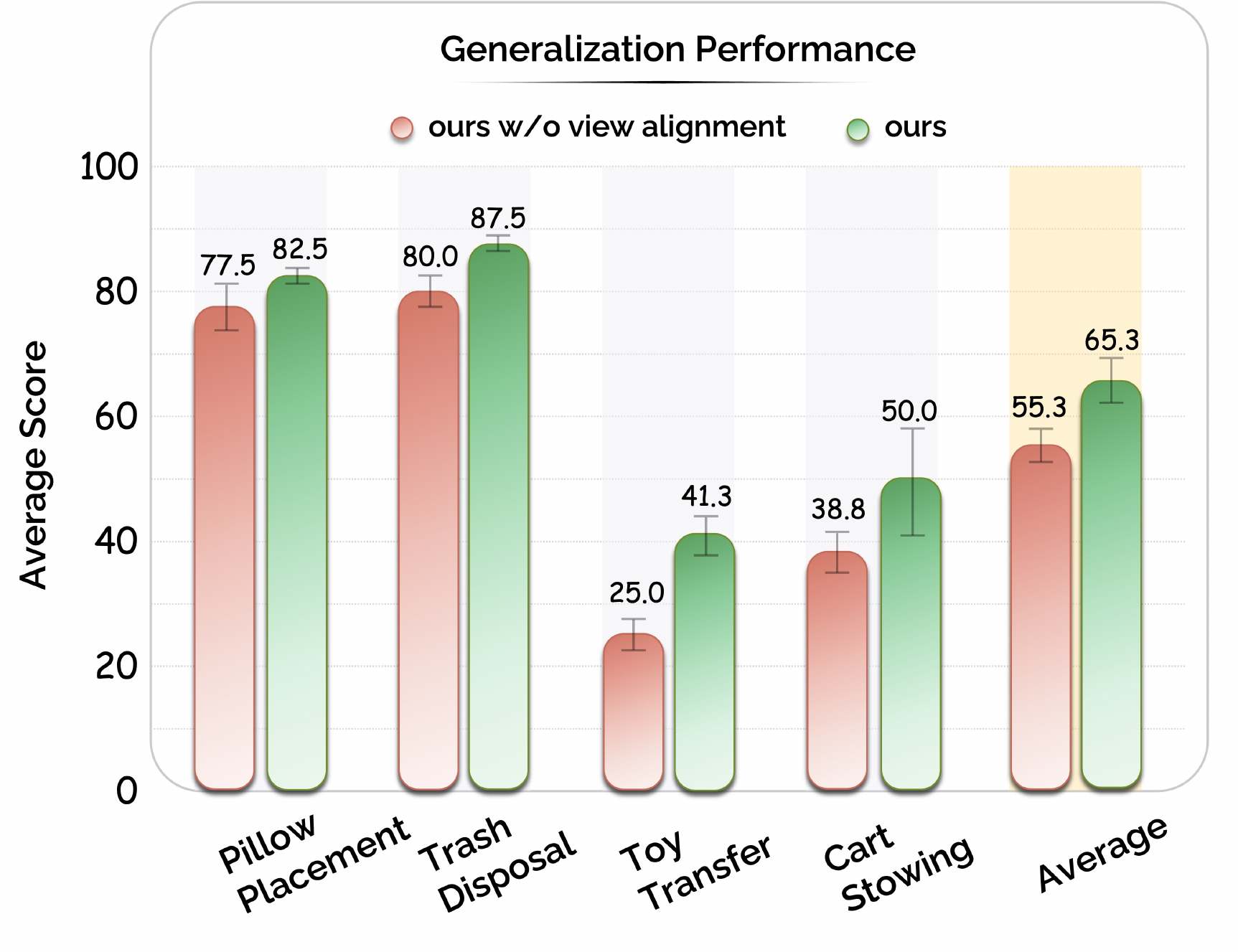}
    \caption{\textbf{View alignment ablation.}
    View alignment is essential for effective human-robot co-training, especially for tasks involving variable object heights such as Toy Transfer.
    }
    \label{fig:ablation-view}
\end{figure}

\subsection{View Alignment Ablation}
\label{sec:exp-modules}

We ablate the effect of view alignment
in \cref{fig:ablation-view}. View alignment yields consistent improvements across all tasks, with the largest gains on Toy Transfer and Cart Stowing. Both tasks involve objects placed at varying heights, creating deployment viewpoints that differ from both data sources: human demonstrations are captured from a higher camera position, while robot data is collected with objects at different heights. Without view alignment, the policy encounters out-of-distribution viewpoints from either source. Our view transformation bridges this gap by reprojecting human observations to the robot's camera height with added pose perturbations, exposing the policy to broader viewpoint variations during training. 
Visualizations are provided in Appendix~\ref{sec:supp-align-vis}.

\subsection{Scene Diversity in Human Data}

\begin{table*}[t!]
\centering
\caption{\textbf{Effect of human demonstration scene diversity on zero-shot generalization.} We evaluate the Trash Disposal task in a novel scene unseen during training, progressively increasing the number of distinct human demonstration scenes while keeping robot data fixed. Sub-steps \textcolor{LocoText}{s1} (locomotion) and \textcolor{ManiText}{s2} (manipulation) are reported alongside the average task score.}
\label{tab:scene_diversity}
\setlength{\tabcolsep}{10pt}
\renewcommand{\arraystretch}{1.15}
\begin{tabular}{c c c c c}
\toprule
\textbf{Training Data} & \textbf{\# Human Scenes} & \textcolor{LocoText}{\textbf{s1}} & \textcolor{ManiText}{\textbf{s2}} & \textbf{Average Score} \\
\midrule
Robot-only    & 0 & 70  & 45 & 57.5 \\
Co-training   & 1 & 90  & 60 & 75.0 \\
Co-training   & 2 & \textbf{100} & 50 & 75.0 \\
Co-training   & 3 & \textbf{100} & \textbf{65} & \textbf{82.5} \\
\bottomrule
\end{tabular}
\end{table*}

We further investigate how scene diversity affects co-training. Fixing the data quantity at 100 robot and 300 human demonstrations, we vary only the number of distinct human scenes and evaluate zero-shot in a novel scene unseen during training (Table~\ref{tab:scene_diversity}). The score rises monotonically from 57.5\% (robot-only) to 82.5\% with three scenes, indicating that scene diversity is a key driver of generalization even at fixed data quantity.

\subsection{Data Collection Efficiency}

As shown in Table~\ref{tab:collection_time}, human demonstrations are roughly $2\times$ faster than robot teleoperation across all four tasks, reducing the per-episode collection time from 62.1\,s to 39.7\,s on average.

\begin{table*}[t!]
\centering
\caption{	\textbf{Comparison of data collection efficiency.} The average time ($s$) for collection of each data episode shows the preeminent efficiency ($\sim2\times$) of gathering human demonstrations over teleoperating robots.
}
\label{tab:collection_time}
\begin{tabular}{lccccc}
\toprule
\textbf{Method} & \textbf{Pillow Placement} & \textbf{Trash Disposal} & \textbf{Toy Transfer} & \textbf{Cart Stowing} & \textbf{Average} \\
\midrule
Robot teleop & 35.7 & 43.1 & 66.2 & 103.5 & 62.1 \\
Human demo  & 16.2 & 18.3 & 34.5 &  89.9 & \textbf{39.7} \\
\bottomrule
\end{tabular}
\end{table*}
\section{Discussions}
\label{sec:discussion}

Our work demonstrates inspiring pioneering results of learning humanoid loco-manipulation from human data, boosting robot-only baselines by a large margin as shown in \cref{fig:histogram}. Yet, the overall success rate is not satisfactory, and typical failure cases are characterized in \cref{fig:failure}. Due to the instability of the humanoid and the huge embodiment gap, we find a bunch of critical factors that affect the performance significantly, even when equipped with the proposed pipeline. We present the devil in the details below, and also incorporate practical guidelines for data collection in Appendix~\ref{sec:supp-guideline}.

\boldparagraph{Action Representation} Two candidate action spaces exist for human-robot alignment: absolute end-effector pose in camera frame and delta end-effector pose. We adopt the latter because human and humanoid morphologies yield systematically different hand-to-camera distances—human arms are typically longer relative to torso height, resulting in different absolute position distributions even for functionally equivalent poses. Delta end-effector naturally shares an identical action space.

However, delta end-effector poses present a fundamental limitation: the correspondence between human hand orientation and robot gripper orientation is ambiguous. While we define explicit pose correspondences, these mappings are difficult to infer from visual observations alone. Consequently, without proprioceptive input, precise rotation transfer remains challenging—the policy cannot reliably disambiguate intended end-effector orientations from egocentric images, limiting fine-grained rotational control in manipulation tasks.

\boldparagraph{Data Scale Requirements} Humanoid loco-manipulation presents unique data efficiency challenges compared to fixed-base manipulation. These tasks require the robot to first navigate to the target location before performing manipulation—yet the robot rarely stops at precisely the same position, resulting in highly diverse viewpoints during the manipulation phase. This viewpoint variability 
raises substantially increasing data requirements. In our experiments, achieving comparable success rates on loco-manipulation tasks requires approximately 2–3× more demonstrations than fixed-base manipulation tasks of similar difficulty, highlighting the compounded challenge of learning both robust navigation and precise manipulation from egocentric observations.

\section{Conclusion} 
\label{sec:conclusion}

We introduce \modelname, the first framework to realize human-to-humanoid transfer for loco-manipulation. By aligning egocentric human demonstrations with robot data through view transformation and unified action space, our approach enables effective VLA co-training. Experiments demonstrate substantial generalization improvements, enabling deployment in diverse environments without in-the-wild robot data. We hope this work encourages broader exploration of egocentric human data toward generalizable humanoid control.

\boldparagraph{Limitations and Future Work} 
As an early endeavor, our method still bears limitations with respect to orientation ambiguity in delta end-effector representations, scaling laws with human data, and expressive whole-body controls. We will take scaling up this paradigm with advanced egocentric hardware as  key future direction.
More discussions are in Appendix~\ref{sec:supp-questions}.

\section*{Acknowledgments}
This study is supported by National Natural Science Foundation of China (62206172). This work is in part supported by the JC STEM Lab of Autonomous Intelligent Systems funded by The Hong Kong Jockey Club Charities Trust. We gratefully acknowledge our robot operators for data collection and evaluations. We also extend our thanks to Jiazhi Yang, Tianyu Li, Yixuan Pan, Junli Ren, Penglin Fu, and Kaiyang Wu for their insightful feedback and fruitful discussions.

{
\bibliographystyle{plainnat}
\bibliography{bibliography_short, bibliography_custom}
}

\clearpage
\newpage
\onecolumn

\appendix

Demo recordings are provided along with this supplementary material. 
We will open-source our code and data to facilitate future research in learning humanoid policies with human demonstrations.

In the appendix, we first provide a list of motivating questions in Sec.~\ref{sec:supp-questions}, and practical guidelines for data collection to benefit human-to-robot transfer in Sec.~\ref{sec:supp-guideline}. Implementation details on \modelname and the real-world experiments are presented in Sec.~\ref{sec:supp-implementation}. Sec.~\ref{sec:supp-score} outlines the scoring criteria when evaluating the experiments to get the scores in the main paper. We show qualitative results on human-to-humanoid alignment in Sec.~\ref{sec:supp-align-vis}. Lastly, the licenses of the assets used in our work are provided in Sec.~\ref{sec:supp-license}.

\subsection{Motivating Questions}
\label{sec:supp-questions}
To help an instinctive and thorough comprehension of our work, we present several motivating questions that might be raised and corresponding answers below.

\bigskip
\noindent\textbf{Q1.} \textit{Why co-train two sources of data in the post-training stage? How about leveraging human data in pre-training?}

\smallskip
These two research focuses are compatible and can work harmoniously. Pre-training with human data has been a hot topic in recent years~\cite{zhu2025emma, cai2025human0}, especially in enhancing the generality of the foundation model, while effective co-training embodied policies remains puzzled.
Co-training in post-training serves as an initial investigation into leveraging egocentric human data for humanoid loco-manipulation, while verifying that our alignment pipeline successfully bridges human and robot data. The successful in-the-wild deployment demonstrates that scene generalization through human data is achievable, suggesting a scalable recipe for future pre-training: collect modest robot demonstrations per task in laboratories, then augment with large-scale human demonstrations across diverse scenarios.

\bigskip
\noindent\textbf{Q2.} \textit{How will \modelname~apply to other humanoid embodiments, besides the mid-sized humanoid robot, Unitree G1?}

\smallskip
Our pipeline addresses alignment between human demonstrations and humanoid action representations, \textit{e.g.}, delta end-effector poses and discrete locomotion commands, which can be instantiated on any humanoid with appropriate inverse kinematics and locomotion controllers. The successful deployment on the mid-sized G1 (1.3m) validates our approach under challenging conditions with a substantial embodiment gap compared to human demonstrators (1.6–1.8m). Full-sized humanoids would exhibit smaller gaps, potentially making transfer more straightforward with minimal modifications.

\bigskip
\noindent\textbf{Q3.} \textit{Can this paradigm be extended to whole-body humanoid VLA with expressive motions, beyond the presented loco-manipulation tasks?}

\smallskip
Our implementation builds upon GR00T-WholeBodyControl~\cite{GR00T-WholeBodyControl}, which constrains lower-body motion to coarse commands for stability. Incorporating more expressive locomotion policies would extend our framework to tasks requiring precise leg movements, such as stepping on pedal-operated trash bins or navigating narrow gaps. The human demonstration pipeline already captures full-body motion, so the data infrastructure supports such extensions, and we envision this as a natural direction for future development. The primary challenge lies in developing robust low-level controllers for expressive whole-body motions while maintaining balance~\cite{luo2025sonic, omnih2o, twist, pan2025ams}.

\bigskip
\noindent\textbf{Q4.} \textit{Why use the egocentric view only, without the commonly adopted wrist cameras? Why not inpaint human hands with robot grippers?}

\smallskip
We adopt a minimalist sensor configuration to establish a clean baseline for human-to-humanoid transfer. Wrist cameras would further amplify the embodiment gap through differing hand geometry and interaction patterns, warranting dedicated investigation~\cite{kareer2025emergence}. Hand inpainting~\cite{lepert2025phantom, li2025h2r, lepert2025masquerade}, while potentially reducing the visual domain gap, introduces errors from imperfect detection and rendering, and is embodiment-specific rather than agnostic. Our pipeline maintains generality through viewpoint transformation and action alignment without tremendous appearance modification. Wrist camera integration and inpainting remain promising future directions.

\bigskip
\noindent\textbf{Q5.} \textit{Limitations and future endeavors.}

\smallskip
Key limitations include: (1) delta end-effector poses create rotation ambiguity without proprioceptive input, limiting fine-grained rotational control; (2) loco-manipulation requires more demonstrations than fixed-base tasks due to viewpoint variability from navigation; (3) discrete locomotion commands limit achievable behaviors. Future work will focus on scaling with advanced egocentric hardware, investigating pre-training with internet-scale egocentric video, and developing expressive action representations preserving cross-embodiment transferability. Besides, similar to general methods of learning from human data~\cite{yuan2025motiontrans, kareer2025emergence}, several practical human demonstration collection guidelines are necessary for efficient co-training, which are listed in Sec.~\ref{sec:supp-guideline}. Random human motion with occluded hands or high-speed motion changes would require more advanced sensor setups, such as tactile and SLAM techniques, to compensate for the estimation error.

\bigskip
\noindent\textbf{Q6.} \textit{Broader impact.}

\smallskip
On the technical side, our alignment between human and humanoid data establishes a form of cross-embodiment adaptation, suggesting humanoid-embodiment-agnostic learning is feasible. 
It hints at the potential for foundation models that learn from human demonstrations and deploy across diverse platforms with minimal adaptation, besides humanoid robots, mobile manipulators, and other non-anthropomorphic platforms. 
For societal impact, our work could motivate large-scale human data collection, which may accelerate innovation and broaden participation across institutions and regions. It can also democratize humanoid development for groups without expensive hardware, though it raises considerations around data privacy and contributor recognition as the paradigm scales.

\subsection{Practical Guidelines for Data Collection}
\label{sec:supp-guideline}

Due to the huge embodiment gap between humans and robots, several protocols are necessary for the demonstrators to follow to realize effective and efficient co-training, especially when we are experimenting with mid-sized humanoids. Though these guidelines may limit human motions to certain extents, they do not constrain the scenario diversity or affect collection efficiency, which are the main advantages of in-the-wild human data.

\boldparagraph{Hand-Wrist Pose Consistency} Demonstrators should maintain consistent hand-wrist orientation patterns throughout task execution. Since our action representation uses delta end-effector poses, small wrist rotations can produce dramatically different action labels even for visually similar manipulation scenarios. Such inconsistencies introduce significant noise into the training data, as nearly identical visual observations become paired with conflicting action targets, ultimately degrading policy learning effectiveness.

\boldparagraph{Base Coordinate Stability} Our pipeline computes locomotion commands using the human waist pose as the base coordinate frame. Consequently, demonstrators should maintain a relatively stable torso posture during movement. Excessive upper body swaying, which is common during natural human walking but absent in humanoid locomotion, can introduce systematic noise in velocity estimation, creating a mismatch between demonstrated and executable locomotion commands.

\boldparagraph{Hand Visibility Maintenance} Reliable hand pose estimation requires minimizing occlusion during manipulation, particularly when ego-centric (head) cameras are used only without wrist cameras. When hands are obscured by manipulated objects or environmental elements, tracking quality degrades substantially, resulting in missing or inaccurate action labels. Demonstrators should adjust their approach angles and manipulation strategies to maintain clear hand visibility throughout task execution, even when this deviates from the most natural human behavior.

\subsection{Implementation Details}
\label{sec:supp-implementation}

\subsubsection{\modelname}
\label{sec:supp-method-detail}

\paragraph{View Alignment}
We employ MoGe~\cite{wang2025moge} at a resolution of $720 \times 1280$ for depth prediction, followed by bilinear upsampling to match the input image resolution.
A rigid camera transformation simulates downward movement from the human to robot viewpoint, with a translation distance of $0.25$\,m.
During training, we apply data augmentation by adding uniform noise of $\pm 0.05$\,m to the translation distance.
The resulting warped images with occlusions are inpainted using Stable Diffusion 2.0 Inpainting~\cite{Rombach_2022_CVPR} with $20$ denoising steps and classifier-free guidance scale of $7.5$.

\paragraph{Action Alignment}
The Savitzky--Golay filter for upper-body translation smoothing uses a window size of 11 and polynomial order 3. The same configuration is applied for rotation smoothing in the SO(3) tangent space and for lower-body pelvis trajectory smoothing.
For lower-body command quantization, forward/backward velocity is discretized into 3 bins, lateral velocity into 3 bins, and yaw rate into 3 bins. The pelvis height change (delta\_height) is computed as the inter-frame difference of the downsampled z-coordinates, where the first frame is set to 0 and subsequent frames represent the change relative to the previous frame.

\subsubsection{Experiments}
\label{sec:supp-exp-detail}

\paragraph{Policy Architecture and Training}
We finetune $\pi_{0.5}$~\cite{intelligence2025pi05} with the following configuration.
The input RGB image is resized to $224 \times 224$.
The output action space is 18-dimensional, consisting of 12-DoF delta end-effector pose (6-DoF $\times$ 2 arms), 3-dimensional discrete locomotion commands (navigation in x, y, and yaw), 2-dimensional binary gripper commands (one per hand), and 1-dimensional delta height command.
The action chunk size is 50 steps.

Training is conducted on 8 $\times$ NVIDIA A100 GPUs with a per-GPU batch size of 32, yielding an effective batch size of 256.
We use the AdamW optimizer with a learning rate of $5 \times 10^{-5}$, weight decay of $1 \times 10^{-10}$ (effectively negligible), gradient clipping norm of 1.0, and a cosine decay learning rate schedule with 10{,}000 warmup steps.
Training runs for 20{,}000 steps.
We use mixed-precision (bfloat16) training throughout.
No proprioceptive inputs are used for either data source.

\paragraph{Robot Platform}
The ZED X Mini camera is mounted on the robot head and captures egocentric RGB images at 20\,Hz with a resolution of $960 \times 540$, which are later downsampled to $224 \times 224$ for model input.
The low-level whole-body controller~\cite{GR00T-WholeBodyControl} converts high-level delta end-effector commands to joint-level motor targets.
Robot demonstration data is collected at 100\,Hz and downsampled to 20\,Hz (factor of 5) for training, with Savitzky-Golay filtering applied to smooth end-effector trajectories.

\paragraph{Policy Inference}
The policy generates 50-step action chunks at each inference.
The inference client runs on RTX 4090, with each forward pass taking approximately 130 ms.
We employ a temporal smoothing mechanism based on linear blending of action chunks.
When a new chunk arrives, we drop the first $k$ actions that has been executed during inference for latency compensation, then linearly blend the overlapping region with the remaining actions from the previous chunk using weights $w_{\text{old}}(i) = 1 - i/(m-1)$ and $w_{\text{new}}(i) = i/(m-1)$ for $i \in [0, m)$, where $m$ is the overlap length.
This ensures smooth transitions between chunks while maintaining responsiveness.

\subsection{Scoring Criteria for Real-world Experiments}
\label{sec:supp-score}
In this section, we introduce four real-world loco--manipulation tasks and the criteria for determining success or failure.
Each task is decomposed into an ordered sequence of stages $\{S_k\}$, where each stage is labeled as either locomotion-dominant or manipulation-dominant.
A trial is considered \textbf{successful} only if \textbf{all} stages succeed in order; otherwise, it is a \textbf{failure}. We evaluate each task in both in-domain and generalized environments.
We additionally record \textbf{stage-wise success/failure} to enable fine-grained analysis.

For consistency with Figs.~\textcolor{red}{5, 7, 8} in the main paper, we also report an \textit{Average Score} for each task, defined as the average stage completion rate.
Specifically, for a task with $K$ ordered stages, each stage is assigned a binary indicator $s_k \in \{0,1\}$, where $s_k = 1$ if stage $S_k$ succeeds and $s_k = 0$ otherwise.
The per-trial task score is defined as
\[
\text{Score} = \frac{1}{K} \sum_{k=1}^{K} s_k .
\]

The \textit{Average Score} reported in Figs.~\textcolor{red}{5, 7, 8} is obtained by averaging $\text{Score}$ over trials, which is equivalent to the mean of the per-stage success rates within the task.

\begin{itemize}[leftmargin=*]

    \item \textbf{Pillow Placement.}
    
    \textit{Task description:}
    the robot starts while grasping a pillow and facing a bed. It must keep holding the pillow, walk to the bed, squat to an appropriate height, and place the pillow onto the target area near the head of the bed. This task is naturally split into a locomotion stage (carrying without dropping) and a placement stage.

    \textit{Experimental setup:} the in-domain environment is a laboratory setting, where the bed is a portable camp bed. In the generalized environment, we recreate a home-like bedroom scene with a standard household bed and consistent background furniture and decor, matching everyday living conditions.

    \textit{Stage-wise success/failure:}
    \begin{itemize}[leftmargin=*]
        \item \textbf{S1 (Locomotion).}
        \begin{itemize}[leftmargin=1.2em, itemsep=0.15em]
            \item \textbf{Success:} the robot reaches a position in front of the bed suitable for subsequent placement \emph{while continuously holding} the pillow (no drop, no unintended release).
            \item \textbf{Failure:} the pillow is dropped/released or the robot stops in a pose/location that makes the next stage infeasible (\textit{e.g.}, too far to reach the target placement region).
        \end{itemize}
        \item \textbf{S2 (Manipulation).}
        \begin{itemize}[leftmargin=1.2em, itemsep=0.15em]
            \item \textbf{Success:} after squatting, the robot places the pillow onto the designated head-area region of the bed.
            \item \textbf{Failure:} the pillow is placed outside the target region, slips/falls off the bed, is released prematurely, or the robot cannot complete the squat-and-place motion.
        \end{itemize}
    \end{itemize}

    \item \textbf{Trash Disposal.}
    
    \textit{Task description:}
    the robot starts while grasping a piece of trash and facing a waste container. It must keep holding the trash, walk to the container, and dispose the trash into the container opening.

    \textit{Experimental setup:} in-domain trials are conducted in the laboratory. For generalized trials, we relocate the waste container to an outdoor garden setting and evaluate the same disposal procedure under the resulting scene and background changes.

    \textit{Stage-wise success/failure:}
    \begin{itemize}[leftmargin=*]
        \item \textbf{S1 (Locomotion).}
        \begin{itemize}[leftmargin=1.2em, itemsep=0.15em]
            \item \textbf{Success:} the robot approaches the waste container to a feasible disposal pose while maintaining a stable grasp on the trash.
            \item \textbf{Failure:} the trash is dropped/released, the robot falls/becomes unstable, or the final approach pose is not suitable for the disposal motion.
        \end{itemize}
        \item \textbf{S2 (Manipulation).}
        \begin{itemize}[leftmargin=1.2em, itemsep=0.15em]
            \item \textbf{Success:} the trash ends up inside the waste container after the disposal attempt (successful insertion through the opening) and does not remain outside or get stuck at the rim as an external object.
            \item \textbf{Failure:} the trash misses the opening, stays outside the container, gets stuck externally at the rim region, or the robot cannot execute the disposal motion stably.
        \end{itemize}
    \end{itemize}

    \item \textbf{Toy Transfer.}
    
    \textit{Task description:}
    a toy is placed on a small elevated stand in front of the robot, and a table is located at a distance. The robot must walk to the stand, grasp the toy with both hands, turn and walk to the table while holding the toy, and finally place the toy onto the table.

    \textit{Experimental setup:}
    in-domain trials are conducted in the laboratory. For generalized trials, we replace the lab desk with household counterparts—using a secretaire as the table and a home cabinet as the stand—and evaluate the task in a home-like setting with consistent background furniture and decor.
    
    \textit{Stage-wise success/failure:}
    \begin{itemize}[leftmargin=*]
        \item \textbf{S1 (Locomotion).}
        \begin{itemize}[leftmargin=1.2em, itemsep=0.15em]
            \item \textbf{Success:} the robot reaches a pose where the toy is reachable for a two-hand grasp attempt.
            \item \textbf{Failure:} the robot stops too far / in a poor pose that makes grasping infeasible.
        \end{itemize}
        \item \textbf{S2 (Manipulation).}
        \begin{itemize}[leftmargin=1.2em, itemsep=0.15em]
            \item \textbf{Success:} the robot lifts and secures the toy with both hands, and the toy remains firmly held (no drop) at the end of the grasp stage.
            \item \textbf{Failure:} failure to establish a stable grasp, dropping the toy during/after lift, or grasping in a way that prevents carrying to the next stage.
        \end{itemize}
        \item \textbf{S3 (Locomotion).}
        \begin{itemize}[leftmargin=1.2em, itemsep=0.15em]
            \item \textbf{Success:} the robot turns and walks to the table while constantly holding the toy, ending at a pose suitable for placement.
            \item \textbf{Failure:} dropping/releasing the toy or arriving at an infeasible placement pose.
        \end{itemize}
        \item \textbf{S4 (Manipulation).}
        \begin{itemize}[leftmargin=1.2em, itemsep=0.15em]
            \item \textbf{Success:} the robot places the toy onto the table surface and the toy remains stably on the table after release.
            \item \textbf{Failure:} the toy is dropped off the table, placed unstably (immediately falls), or the robot cannot complete the placement.
        \end{itemize}
    \end{itemize}

    \item \textbf{Cart Stowing.}
    
    \textit{Task description:}
    a toy is placed on a table, and a cart is located nearby. The robot must first hold the cart handle and move the cart to a position beside the table such that the toy becomes reachable by the left hand; then it releases the handle, grasps the toy with the left hand, moves the toy over the cart basket and releases it into the cart; finally it pushes the cart forward. This task includes sustained cart contact during locomotion and multi-phase manipulation.

    \textit{Experimental setup:} in-domain trials are conducted in the laboratory. For generalized trials, we replace the table with a retail-style shelf and place the toy on the shelf to better match a shopping scenario. The generalized evaluation is performed in a real convenience store with natural background clutter and layout.

    \textit{Stage-wise success/failure:}
    \begin{itemize}[leftmargin=*]
        \item \textbf{S1 (Locomotion).}
        \begin{itemize}[leftmargin=1.2em, itemsep=0.15em]
            \item \textbf{Success:} the robot maintains a stable grasp on the cart handle and pushes/moves the cart to a position where the toy on the table is reachable by the left hand.
            \item \textbf{Failure:} stopping in a pose that makes toy grasping infeasible.
        \end{itemize}
        \item \textbf{S2 (Manipulation).}
        \begin{itemize}[leftmargin=1.2em, itemsep=0.15em]
            \item \textbf{Success:} the robot releases the handle and establishes a stable left-hand grasp on the toy.
            \item \textbf{Failure:} failed grasp, dropping the toy, or grasping in a way that prevents transporting it over the cart.
        \end{itemize}
        \item \textbf{S3 (Manipulation).}
        \begin{itemize}[leftmargin=1.2em, itemsep=0.15em]
            \item \textbf{Success:} while holding the toy, the robot moves the left hand above the cart basket and releases the toy such that it lands inside the cart.
            \item \textbf{Failure:} the toy lands outside the cart, gets stuck on the rim/edge externally, is dropped prematurely, or the robot cannot execute the transfer motion stably.
        \end{itemize}
        \item \textbf{S4 (Locomotion).}
        \begin{itemize}[leftmargin=1.2em, itemsep=0.15em]
            \item \textbf{Success:} the robot pushes/moves the cart forward in a controlled and stable manner after the toy has been stowed.
            \item \textbf{Failure:} unstable pushing leading to loss of control or fall, or requiring human intervention.
        \end{itemize}
    \end{itemize}

\end{itemize}

\subsection{Qualitative Results on Human-to-Humanoid Alignment}
\label{sec:supp-align-vis}

Fig.~\ref{fig:supp-alignment-vis} visualizes the view alignment pipeline across all four tasks. For each example, we show the original human egocentric image, the estimated depth map from MoGe~\cite{wang2025moge}, the reprojected image after point cloud transformation to the robot camera frame, and the final image after diffusion-based inpainting. The reprojected images exhibit a noticeable downward viewpoint shift consistent with the lower robot camera height, introducing black regions from disocclusions and invalid depth predictions. The inpainting step faithfully fills these missing regions while preserving scene structure and object appearance, producing complete RGB images that closely approximate what the robot would observe from its egocentric viewpoint. Notably, the pipeline handles diverse scenes including indoor bedrooms, outdoor gardens, cluttered tabletops, and retail environments, demonstrating its generality across the tasks and scenarios used in our experiments.

\begin{figure*}[t!]
    \centering
    \includegraphics[width=\linewidth]{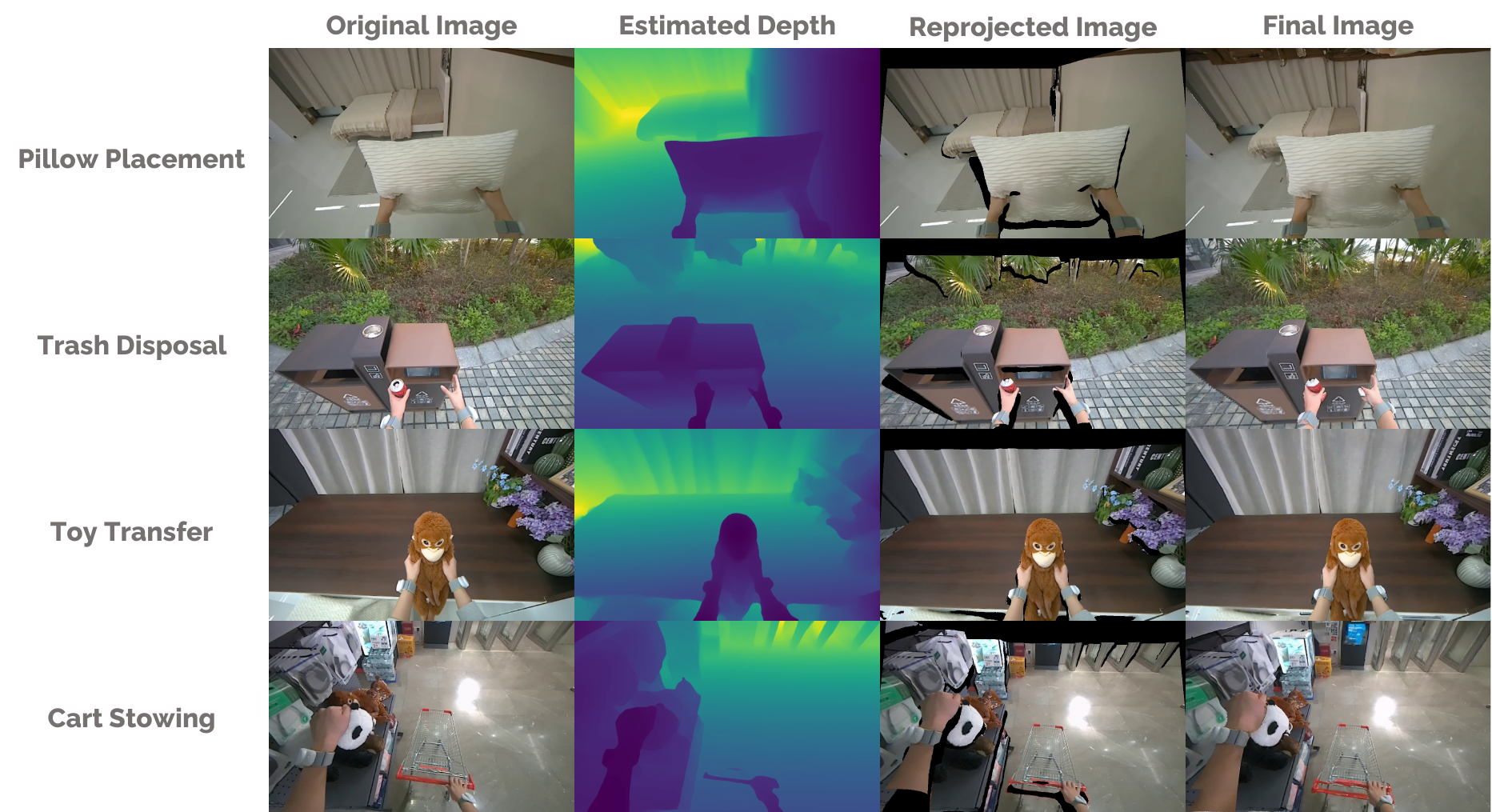}
    \caption{\textbf{Qualitative results of the view alignment pipeline.} For each task, we show the original human egocentric image, the MoGe-estimated depth map, the reprojected image after point cloud transformation to the robot camera frame, and the final inpainted result. The pipeline produces a consistent downward viewpoint shift that approximates the robot's lower camera height, with inpainting filling disoccluded regions to yield complete RGB observations.}
    \label{fig:supp-alignment-vis}
\end{figure*}

\subsection{License of Assets}
\label{sec:supp-license}

This work builds upon several open-source projects with permissive licenses. Our implementation incorporates PICO XRoboToolkit~\cite{zhao2025xrobotoolkit}, MoGe~\cite{wang2025moge}, and latent diffusion~\cite{Rombach_2022_CVPR}, all released under the MIT License. The MoGe codebase additionally includes components from DINOv2~\cite{oquab2023dinov2}, which are released by Meta AI under the Apache 2.0 License. We also make use of openpi~\cite{intelligence2025pi05}, which is distributed under the Apache 2.0 License. Finally, our system integrates GR00T-WholeBodyControl~\cite{GR00T-WholeBodyControl}, which is released under the NVIDIA non-commercial license. We will open-source our assets, including code and models, under the Apache 2.0 License.

\end{document}